\documentclass[sigconf]{aamas}

\usepackage{balance} 

\makeatletter
\gdef\@copyrightpermission{
  \begin{minipage}{0.2\columnwidth}
   \href{https://creativecommons.org/licenses/by/4.0/}{\includegraphics[width=0.90\textwidth]{by}}
  \end{minipage}\hfill
  \begin{minipage}{0.8\columnwidth}
   \href{https://creativecommons.org/licenses/by/4.0/}{This work is licensed under a Creative Commons Attribution International 4.0 License.}
  \end{minipage}
  \vspace{5pt}
}
\makeatother

\setcopyright{ifaamas}
\acmConference[AAMAS '25]{Proc.\@ of the 24th International Conference
on Autonomous Agents and Multiagent Systems (AAMAS 2025)}{May 19 -- 23, 2025}
{Detroit, Michigan, USA}{Y.~Vorobeychik, S.~Das, A.~Nowé  (eds.)}
\copyrightyear{2025}
\acmYear{2025}
\acmDOI{}
\acmPrice{}
\acmISBN{}

\acmSubmissionID{ea0109}

\title{Predicting Team Performance from Communications in Simulated Search-and-Rescue}

\subtitle{Extended Abstract}

\author{Ali Jalal-Kamali}
\affiliation{
  \institution{University of Southern California}
  \city{Los Angeles}
  \country{USA}}
\email{jalalkam@usc.edu}

\author{Nikolos M. Gurney}
\affiliation{
  \institution{University of Southern California}
  \city{Los Angeles}
  \country{USA}}
\email{gurney@ict.usc.edu}

\author{David V. Pynadath}
\affiliation{
  \institution{Rice University}
  \city{Houston}
  \country{USA}}
\email{pynadath@rice.edu}

\begin{abstract}
Understanding how individual traits influence team performance is valuable, but these traits are not always directly observable. Prior research has inferred traits like trust from behavioral data. We analyze conversational data to identify team traits and their correlation with teaming outcomes. Using transcripts from a Minecraft-based search-and-rescue experiment, we apply topic modeling and clustering to uncover key interaction patterns. Our findings show that variations in teaming outcomes can be explained through these inferences, with different levels of predictive power derived from individual traits and team dynamics.
\end{abstract}


\newcommand{\BibTeX}{\rm B\kern-.05em{\sc i\kern-.025em b}\kern-.08em\TeX}

\begin{document}


\pagestyle{fancy}
\fancyhead{}


\maketitle 


\section{Introduction}

Autonomous agents have begun leveraging artificial intelligence to improve human teamwork through automated assessment and assistance during task performance \cite{seo2021towards,sukthankar2007towards,webber2019team}. However, their usefulness is limited by an agent's ability to understand the people it wants to help. When interacting with a \textit{team} of humans, the agent must model not only the multiple individuals but also their relationships and interactions. When multiple people work together, communication becomes integral to their behavior across domains \cite{emmitt2006communication, marlow2017communication, stempfle2002thinking}. This communication provides valuable information about team characteristics and performance \cite{tiferes2018impact}. However, the value depends on the agent's ability to understand this naturally occurring communication, which isn't oriented toward AI systems.


\section{Experimental Testbed Data}\label{sec:data}
We analyze data from Study 3 of DARPA's Artificial Social Intelligence for Successful Teams (ASIST) program \cite{ASISTStudy3}. The study used a Minecraft-based urban search and rescue task \cite{freeman2021evaluating,corral2021building,johnson2016malmo} where teams of three participants worked together. Team members had distinct roles: the \textit{medic} treats victims, the \textit{engineer} clears obstacles, and the \textit{transporter} efficiently moves victims.

We use the following components in our study

\begin{itemize}
    \item Communication transcripts: teams communicated via audio, which was transcribed. Our analysis uses only these transcripts, ignoring simulation logs and AI advisor data.
    \item Pre-trial team profiles: The dataset includes eight Background of Experience, Affect, and Resources Diagnostic (BEARD) variables such as anger, anxiety, etc. The BEARD variables measure the team characteristics before trials.
    \item Dynamic effectiveness Diagnostic (TED) measures: The ASIST testbed also contains variables that measure different aspects of team effectiveness throughout the trials, without any interactions with the team. The TED variables generate a dynamic stream of team processes measures.
\end{itemize}


\section{Methodology and Results}\label{sec:method}
Our analysis aims to identify teams needing intervention and determine how early to intervene based on the communication patterns.

\subsection{BEARD Profiles}
Linear regression of BEARD variables against performance revealed several significant relationships
\begin{itemize}
    \item anger: showed a strong negative correlation.
    \item social perceptiveness: demonstrated a positive correlation.
    \item transporting skill: showed an unexpected negative correlation, though this may be due to overconfidence effects.
\end{itemize}

\subsection{TED Measures}
Once we had the impact of the pre-trial variables, we turned our focus to the variables that measure different aspects of team effectiveness throughout the trials, without any interactions with the team members.
Since some of these measures inform other ones, e.g., process-effort-s informs process-effort-agg, to exclude all inter-dependencies of TED variables, we only included variables that are aggregates, time measures, and communication-based. 

A linear regression of TED variables and scores indicates
\begin{itemize}
    \item process-effort-agg: has a positive coefficient.
    \item comms-total-words: has a positive coefficient.
    \item process-skill-use-agg: has a negative coefficient.
\end{itemize}

\subsection{Pre-Processing of Transcripts}
Each transcript contains text for a complete experimental session with teams performing the task twice (two trials). We removed administrative text, split files into individual trials, and eliminated redundant trials, resulting in 222 unique trial transcripts. We created document-term matrices using standard preprocessing: lowercase conversion, punctuation/number removal, and stopword filtering.

\subsection{Intra-team Communication Analysis}
After pre-processing, we investigated the content of the conversations to inform our process about the team's performance. Topic modeling is an unsupervised method of extracting potential topics from the text, where such topics are representative of the main content of a document. But first we need to find the best number of topics.We applied Latent Dirichlet Allocation (LDA) topic modeling using textmineR \cite{c:23}. For each topic count (2-20), we ran LDA 100 times per count, evaluating the average probabilistic coherences; based on which, we selected 12 as the topic count.
Table \ref{table:1} shows the most probable words per topic. While some topics share terms due to the common search-and-rescue context, others (like topics 2, 3, and 5) are distinct, revealing different communication patterns.

\begin{table}[t]
    \caption{Most probable words for the 12 topics.}
    \label{table:1}
    \begin{tabular}{cc} \toprule 
        \textit{Topic} & \textit{Top Most Probable Words} \\ \midrule
        1 & one, critical, see, right, go \\ 
        2 & meeting, one, critical, two, management \\
        3 & need, come, go, yeah, okay \\
        4 & patient, one, critical, engineer, patients \\
        5 & green, blue, hallway, victims, red \\
        6 & yeah, okay, oh, just, right \\
        7 & transporter, engineer, medic, victim, victims \\
        8 & go, one, okay, ahead, think \\
        9 & victim, room, victims, critical, see \\
        10 & engineer, one, just, get, right \\
        11 & critical, victim, engineer, victims, type \\
        12 & right, critical, just, like, going \\ \bottomrule
    \end{tabular}
\end{table}

\subsection{Categorization Abstraction}
In order to have an abstraction over the categories that may be present among the trial conversations and to find potential subgroups that could indicate various performances, we can perform a clustering over the topic probability distributions for the trials using theta matrix as the variables. Using gap statistics, we determined 8 as the optimal number of clusters. K-means clustering revealed strong differentiation between first and second trials (Table \ref{table:2}), despite not using performance data.

\begin{table}[t]
    \caption{Trial-one vs. trial-two separation using clustering}
    \label{table:2}
    \begin{tabular}{ccc}\toprule 
        \textit{Cluster} & \textit{Trial One} & \textit{Trial Two} \\ \midrule
        1 & 29\% & \textbf{71\%} \\ 
        2 & \textbf{84\%} & 16\% \\
        3 & \textbf{84\%} & 16\% \\
        4 & 14\% & \textbf{86\%} \\
        {\color{red} 5} & {\color{red} 53\%} & {\color{red} 47\%} \\
        6 & 11\% & \textbf{89\%} \\
        7 & \textbf{91\%} & 9\% \\
        8 & 29\% & \textbf{71\%} \\ \bottomrule
    \end{tabular}
\end{table}

To investigate how these clusters relate to performance, we examined the score distributions per cluster. Linear regression showed significant relationships between cluster assignment and performance (p=0.0008), indicating a strong relation between cluster assignment and trial outcomes. The analysis revealed 
\begin{itemize}
    \item Cluster 5: Lowest performance (coefficient: -196.30)
    \item Cluster 2: Second-lowest (coefficient: -147.753)
    \item Cluster 3: Third-lowest (coefficient: -115.095)
\end{itemize}

BEARD logistic regression for cluster 5 showed:
\begin{itemize}
    \item Negative correlation with social perceptiveness
    \item Positive correlations with spatial ability and game skills
\end{itemize}

Team Effectiveness Diagnostic (TED) variables revealed distinctive patterns across clusters
\begin{itemize}
    \item Cluster 5 (lowest performing): high inaction rates, low process effort/coverage/triaging, minimal workload distribution, poor communication balance.
    \item Cluster 2: negative correlation with communication equity, moderate process coverage, uneven task distribution.
    \item Cluster 3: negative correlation with process workload, inconsistent team coordination, mixed communication patterns.
\end{itemize}

These patterns suggest that effective teams maintain balanced communication and workload distribution, while struggling teams show more fragmented interaction patterns.

\subsection{Early Prediction and Intervention Pipeline}
For early predictions, we analyzed transcripts portions
\begin{itemize}
    \item with the first 1/10 of each transcript, we can predict which cluster the trial belongs with 47\% accuracy. 
    \item at 1/3 of each transcript, the accuracy reaches 76\%.
\end{itemize}
The analysis and processes above is used in this pipeline for an autonomous agent to identify the low performing trials
\begin{enumerate}
    \item with 10\% of the trial, the agent predicts trial's cluster.
    \item if the cluster is low-performing, the agent uses the BEARD variables to decide to intervene. 
    \item at 30\% of the trial, the agent predicts the trial's cluster again.
    \item if the cluster is still low performing and the TED measures have not improved since the 10\%, the agent intervenes again. 
    \item repeat steps 3-4 at 50\% and 70\% if the team needs more help. 
\end{enumerate}

Our pipeline allows an agent to predict the team performance early on and take appropriate action, which is quite effective to have a system that allows for such predictions to happen as early as 10 to 30 percentage of the trial.



\begin{acks}
Research was sponsored by the Army Research Office and was accomplished under Cooperative Agreement Number W911NF-20-2-0053. The views and conclusions contained in this document are those of the authors and should not be interpreted as representing the official policies, either expressed or implied, of the Army Research Office or the U.S. Government. The U.S. Government is authorized to reproduce and distribute reprints for Government purposes notwithstanding any copyright notation herein.
\end{acks}



\bibliographystyle{ACM-Reference-Format} 
\bibliography{aamas2025}


\begin{thebibliography}{27}


\ifx \showCODEN    \undefined \def \showCODEN     #1{\unskip}     \fi
\ifx \showDOI      \undefined \def \showDOI       #1{#1}\fi
\ifx \showISBNx    \undefined \def \showISBNx     #1{\unskip}     \fi
\ifx \showISBNxiii \undefined \def \showISBNxiii  #1{\unskip}     \fi
\ifx \showISSN     \undefined \def \showISSN      #1{\unskip}     \fi
\ifx \showLCCN     \undefined \def \showLCCN      #1{\unskip}     \fi
\ifx \shownote     \undefined \def \shownote      #1{#1}          \fi
\ifx \showarticletitle \undefined \def \showarticletitle #1{#1}   \fi
\ifx \showURL      \undefined \def \showURL       {\relax}        \fi
\providecommand\bibfield[2]{#2}
\providecommand\bibinfo[2]{#2}
\providecommand\natexlab[1]{#1}
\providecommand\showeprint[2][]{arXiv:#2}

\bibitem[\protect\citeauthoryear{Akash, Reid, and Jain}{Akash et~al\mbox{.}}{2018}]%
        {akash2018adaptive}
\bibfield{author}{\bibinfo{person}{Kumar Akash}, \bibinfo{person}{Tahira Reid}, {and} \bibinfo{person}{Neera Jain}.} \bibinfo{year}{2018}\natexlab{}.
\newblock \showarticletitle{Adaptive probabilistic classification of dynamic processes: A case study on human trust in automation}. In \bibinfo{booktitle}{\emph{2018 Annual American Control Conference (ACC)}}. IEEE, \bibinfo{pages}{246--251}.
\newblock


\bibitem[\protect\citeauthoryear{Albrecht and Stone}{Albrecht and Stone}{2018}]%
        {albrecht2018autonomous}
\bibfield{author}{\bibinfo{person}{Stefano~V Albrecht} {and} \bibinfo{person}{Peter Stone}.} \bibinfo{year}{2018}\natexlab{}.
\newblock \showarticletitle{Autonomous agents modelling other agents: A comprehensive survey and open problems}.
\newblock \bibinfo{journal}{\emph{Artificial Intelligence}}  \bibinfo{volume}{258} (\bibinfo{year}{2018}), \bibinfo{pages}{66--95}.
\newblock


\bibitem[\protect\citeauthoryear{Andrews, Lilly, Srivastava, and Feigh}{Andrews et~al\mbox{.}}{2023}]%
        {andrews2023role}
\bibfield{author}{\bibinfo{person}{Robert~W Andrews}, \bibinfo{person}{J~Mason Lilly}, \bibinfo{person}{Divya Srivastava}, {and} \bibinfo{person}{Karen~M Feigh}.} \bibinfo{year}{2023}\natexlab{}.
\newblock \showarticletitle{The role of shared mental models in human-{AI} teams: a theoretical review}.
\newblock \bibinfo{journal}{\emph{Theoretical Issues in Ergonomics Science}} \bibinfo{volume}{24}, \bibinfo{number}{2} (\bibinfo{year}{2023}), \bibinfo{pages}{129--175}.
\newblock


\bibitem[\protect\citeauthoryear{Baron-Cohen, Wheelwright, Hill, Raste, and Plumb}{Baron-Cohen et~al\mbox{.}}{2001}]%
        {baron2001reading}
\bibfield{author}{\bibinfo{person}{Simon Baron-Cohen}, \bibinfo{person}{Sally Wheelwright}, \bibinfo{person}{Jacqueline Hill}, \bibinfo{person}{Yogini Raste}, {and} \bibinfo{person}{Ian Plumb}.} \bibinfo{year}{2001}\natexlab{}.
\newblock \showarticletitle{The ``{Reading} the {Mind} in the {Eyes}'' Test revised version: A study with normal adults, and adults with {Asperger} syndrome or high-functioning autism}.
\newblock \bibinfo{journal}{\emph{The Journal of Child Psychology and Psychiatry and Allied Disciplines}} \bibinfo{volume}{42}, \bibinfo{number}{2} (\bibinfo{year}{2001}), \bibinfo{pages}{241--251}.
\newblock


\bibitem[\protect\citeauthoryear{Bendell, Williams, Fiore, and Jentsch}{Bendell et~al\mbox{.}}{2021}]%
        {bendell2021towards}
\bibfield{author}{\bibinfo{person}{Rhyse Bendell}, \bibinfo{person}{Jessica Williams}, \bibinfo{person}{Stephen~M Fiore}, {and} \bibinfo{person}{Florian Jentsch}.} \bibinfo{year}{2021}\natexlab{}.
\newblock \showarticletitle{Towards artificial social intelligence: Inherent features, individual differences, mental models, and theory of mind}. In \bibinfo{booktitle}{\emph{International Conference on Applied Human Factors and Ergonomics}}. \bibinfo{pages}{20--28}.
\newblock


\bibitem[\protect\citeauthoryear{Corral, Tatapudi, Buchanan, Huang, and Cooke}{Corral et~al\mbox{.}}{2021}]%
        {corral2021building}
\bibfield{author}{\bibinfo{person}{Christopher~C Corral}, \bibinfo{person}{Keerthi~Shrikar Tatapudi}, \bibinfo{person}{Verica Buchanan}, \bibinfo{person}{Lixiao Huang}, {and} \bibinfo{person}{Nancy~J Cooke}.} \bibinfo{year}{2021}\natexlab{}.
\newblock \showarticletitle{Building a synthetic task environment to support artificial social intelligence research}.
\newblock \bibinfo{journal}{\emph{Proceedings of the Human Factors and Ergonomics Society Annual Meeting}} \bibinfo{volume}{65}, \bibinfo{number}{1} (\bibinfo{year}{2021}), \bibinfo{pages}{660--664}.
\newblock


\bibitem[\protect\citeauthoryear{Emmitt and Gorse}{Emmitt and Gorse}{2006}]%
        {emmitt2006communication}
\bibfield{author}{\bibinfo{person}{Stephen Emmitt} {and} \bibinfo{person}{Christopher Gorse}.} \bibinfo{year}{2006}\natexlab{}.
\newblock \bibinfo{booktitle}{\emph{Communication in construction teams}}.
\newblock \bibinfo{publisher}{Routledge}.
\newblock


\bibitem[\protect\citeauthoryear{Freeman, Huang, Woods, and Cauffman}{Freeman et~al\mbox{.}}{2021}]%
        {freeman2021evaluating}
\bibfield{author}{\bibinfo{person}{Jared~T Freeman}, \bibinfo{person}{Lixiao Huang}, \bibinfo{person}{Matt Woods}, {and} \bibinfo{person}{Stephen~J Cauffman}.} \bibinfo{year}{2021}\natexlab{}.
\newblock \showarticletitle{Evaluating artificial social intelligence in an urban search and rescue task environment}. In \bibinfo{booktitle}{\emph{AAAI Fall Symposium on Computational Theory of Mind for Human-Machine Teams}}.
\newblock


\bibitem[\protect\citeauthoryear{Gurney, Pynadath, and Wang}{Gurney et~al\mbox{.}}{2022}]%
        {gurney2022measuring}
\bibfield{author}{\bibinfo{person}{Nikolos Gurney}, \bibinfo{person}{David~V Pynadath}, {and} \bibinfo{person}{Ning Wang}.} \bibinfo{year}{2022}\natexlab{}.
\newblock \showarticletitle{Measuring and predicting human trust in recommendations from an AI teammate}. In \bibinfo{booktitle}{\emph{International Conference on Human-Computer Interaction}}. Springer, \bibinfo{pages}{22--34}.
\newblock


\bibitem[\protect\citeauthoryear{Hackman and Wageman}{Hackman and Wageman}{2005}]%
        {hackman2005theory}
\bibfield{author}{\bibinfo{person}{J~Richard Hackman} {and} \bibinfo{person}{Ruth Wageman}.} \bibinfo{year}{2005}\natexlab{}.
\newblock \showarticletitle{A theory of team coaching}.
\newblock \bibinfo{journal}{\emph{Academy of Management Review}} \bibinfo{volume}{30}, \bibinfo{number}{2} (\bibinfo{year}{2005}), \bibinfo{pages}{269--287}.
\newblock


\bibitem[\protect\citeauthoryear{Hegarty, Richardson, Montello, Lovelace, and Subbiah}{Hegarty et~al\mbox{.}}{2002}]%
        {hegarty2002development}
\bibfield{author}{\bibinfo{person}{Mary Hegarty}, \bibinfo{person}{Anthony~E Richardson}, \bibinfo{person}{Daniel~R Montello}, \bibinfo{person}{Kristin Lovelace}, {and} \bibinfo{person}{Ilavanil Subbiah}.} \bibinfo{year}{2002}\natexlab{}.
\newblock \showarticletitle{Development of a self-report measure of environmental spatial ability}.
\newblock \bibinfo{journal}{\emph{Intelligence}} \bibinfo{volume}{30}, \bibinfo{number}{5} (\bibinfo{year}{2002}), \bibinfo{pages}{425--447}.
\newblock


\bibitem[\protect\citeauthoryear{Huang, Freeman, Cooke, Colonna-Romano, Wood, Buchanan, and Caufman}{Huang et~al\mbox{.}}{2022}]%
        {ASISTStudy3}
\bibfield{author}{\bibinfo{person}{Lixiao Huang}, \bibinfo{person}{Jared Freeman}, \bibinfo{person}{Nancy Cooke}, \bibinfo{person}{John~``JCR'' Colonna-Romano}, \bibinfo{person}{Matt Wood}, \bibinfo{person}{Verica Buchanan}, {and} \bibinfo{person}{Stephen Caufman}.} \bibinfo{year}{2022}\natexlab{}.
\newblock \bibinfo{title}{{Artificial Social Intelligence for Successful Teams (ASIST) Study 3}}.
\newblock
\newblock
\urldef\tempurl%
\url{https://doi.org/10.48349/ASU/QDQ4MH}
\showDOI{\tempurl}


\bibitem[\protect\citeauthoryear{Johnson, Hofmann, Hutton, and Bignell}{Johnson et~al\mbox{.}}{2016}]%
        {johnson2016malmo}
\bibfield{author}{\bibinfo{person}{Matthew Johnson}, \bibinfo{person}{Katja Hofmann}, \bibinfo{person}{Tim Hutton}, {and} \bibinfo{person}{David Bignell}.} \bibinfo{year}{2016}\natexlab{}.
\newblock \showarticletitle{The Malmo platform for artificial intelligence experimentation}. In \bibinfo{booktitle}{\emph{Proceedings of the International Joint Conference on Artificial Intelligence}}. \bibinfo{pages}{4246--4247}.
\newblock


\bibitem[\protect\citeauthoryear{{Jones, T.}}{{Jones, T.}}{2021}]%
        {c:23}
\bibfield{author}{\bibinfo{person}{{Jones, T.}}} \bibinfo{year}{2021}\natexlab{}.
\newblock \bibinfo{title}{textmineR Package}.
\newblock \bibinfo{howpublished}{\url{https://cran.r-project.org/web/packages/textmineR/index.html}}.
\newblock
\newblock
\shownote{Accessed: 2024-08-10.}


\bibitem[\protect\citeauthoryear{Maese, Diego-Rosell, DeBusk-Lane, and Kress}{Maese et~al\mbox{.}}{2021}]%
        {maese21}
\bibfield{author}{\bibinfo{person}{Ellyn Maese}, \bibinfo{person}{Pablo Diego-Rosell}, \bibinfo{person}{Les DeBusk-Lane}, {and} \bibinfo{person}{Nathan Kress}.} \bibinfo{year}{2021}\natexlab{}.
\newblock \showarticletitle{Development of emergent leadership measurement: Implications for human-machine teams}. In \bibinfo{booktitle}{\emph{AAAI Fall Symposium on Computational Theory of Mind for Human-Machine Teams}}.
\newblock


\bibitem[\protect\citeauthoryear{Marlow, Lacerenza, and Salas}{Marlow et~al\mbox{.}}{2017}]%
        {marlow2017communication}
\bibfield{author}{\bibinfo{person}{Shannon~L Marlow}, \bibinfo{person}{Christina~N Lacerenza}, {and} \bibinfo{person}{Eduardo Salas}.} \bibinfo{year}{2017}\natexlab{}.
\newblock \showarticletitle{Communication in virtual teams: A conceptual framework and research agenda}.
\newblock \bibinfo{journal}{\emph{Human resource management review}} \bibinfo{volume}{27}, \bibinfo{number}{4} (\bibinfo{year}{2017}), \bibinfo{pages}{575--589}.
\newblock


\bibitem[\protect\citeauthoryear{Myers, Ball, Cooke, Freiman, Caisse, Rodgers, Demir, and McNeese}{Myers et~al\mbox{.}}{2018}]%
        {myers2018autonomous}
\bibfield{author}{\bibinfo{person}{Christopher Myers}, \bibinfo{person}{Jerry Ball}, \bibinfo{person}{Nancy Cooke}, \bibinfo{person}{Mary Freiman}, \bibinfo{person}{Michelle Caisse}, \bibinfo{person}{Stuart Rodgers}, \bibinfo{person}{Mustafa Demir}, {and} \bibinfo{person}{Nathan McNeese}.} \bibinfo{year}{2018}\natexlab{}.
\newblock \showarticletitle{Autonomous intelligent agents for team training}.
\newblock \bibinfo{journal}{\emph{IEEE Intelligent Systems}} \bibinfo{volume}{34}, \bibinfo{number}{2} (\bibinfo{year}{2018}), \bibinfo{pages}{3--14}.
\newblock


\bibitem[\protect\citeauthoryear{O’Neill, McNeese, Barron, and Schelble}{O’Neill et~al\mbox{.}}{2022}]%
        {o2022human}
\bibfield{author}{\bibinfo{person}{Thomas O’Neill}, \bibinfo{person}{Nathan McNeese}, \bibinfo{person}{Amy Barron}, {and} \bibinfo{person}{Beau Schelble}.} \bibinfo{year}{2022}\natexlab{}.
\newblock \showarticletitle{Human--autonomy teaming: A review and analysis of the empirical literature}.
\newblock \bibinfo{journal}{\emph{Human factors}} \bibinfo{volume}{64}, \bibinfo{number}{5} (\bibinfo{year}{2022}), \bibinfo{pages}{904--938}.
\newblock


\bibitem[\protect\citeauthoryear{Pynadath, Wang, and Kamireddy}{Pynadath et~al\mbox{.}}{2019}]%
        {pynadath2019markovian}
\bibfield{author}{\bibinfo{person}{David~V Pynadath}, \bibinfo{person}{Ning Wang}, {and} \bibinfo{person}{Sreekar Kamireddy}.} \bibinfo{year}{2019}\natexlab{}.
\newblock \showarticletitle{A {Markovian} method for predicting trust behavior in human-agent interaction}. In \bibinfo{booktitle}{\emph{Proceedings of the International Conference on Human-Agent Interaction}}. \bibinfo{pages}{171--178}.
\newblock


\bibitem[\protect\citeauthoryear{Seo, Kennedy-Metz, Zenati, Shah, Dias, and Unhelkar}{Seo et~al\mbox{.}}{2021}]%
        {seo2021towards}
\bibfield{author}{\bibinfo{person}{Sangwon Seo}, \bibinfo{person}{Lauren~R Kennedy-Metz}, \bibinfo{person}{Marco~A Zenati}, \bibinfo{person}{Julie~A Shah}, \bibinfo{person}{Roger~D Dias}, {and} \bibinfo{person}{Vaibhav~V Unhelkar}.} \bibinfo{year}{2021}\natexlab{}.
\newblock \showarticletitle{Towards an AI coach to infer team mental model alignment in healthcare}. In \bibinfo{booktitle}{\emph{Proceedings of the IEEE Conference on Cognitive and Computational Aspects of Situation Management}}. \bibinfo{pages}{39--44}.
\newblock


\bibitem[\protect\citeauthoryear{{Sievert C.}}{{Sievert C.}}{2018}]%
        {c:24}
\bibfield{author}{\bibinfo{person}{{Sievert C.}}} \bibinfo{year}{2018}\natexlab{}.
\newblock \bibinfo{title}{LDAvis Package}.
\newblock \bibinfo{howpublished}{\url{https://github.com/cpsievert/LDAvis}}.
\newblock
\newblock
\shownote{Accessed: 2024-08-10.}


\bibitem[\protect\citeauthoryear{Stempfle and Badke-Schaub}{Stempfle and Badke-Schaub}{2002}]%
        {stempfle2002thinking}
\bibfield{author}{\bibinfo{person}{Joachim Stempfle} {and} \bibinfo{person}{Petra Badke-Schaub}.} \bibinfo{year}{2002}\natexlab{}.
\newblock \showarticletitle{Thinking in design teams-an analysis of team communication}.
\newblock \bibinfo{journal}{\emph{Design studies}} \bibinfo{volume}{23}, \bibinfo{number}{5} (\bibinfo{year}{2002}), \bibinfo{pages}{473--496}.
\newblock


\bibitem[\protect\citeauthoryear{Sukthankar, Sycara, Giampapa, Burnett, and Preece}{Sukthankar et~al\mbox{.}}{2007}]%
        {sukthankar2007towards}
\bibfield{author}{\bibinfo{person}{Gita Sukthankar}, \bibinfo{person}{Katia Sycara}, \bibinfo{person}{Joseph~A Giampapa}, \bibinfo{person}{Chris Burnett}, {and} \bibinfo{person}{Alun Preece}.} \bibinfo{year}{2007}\natexlab{}.
\newblock \showarticletitle{Towards a model of agent-assisted team search}. In \bibinfo{booktitle}{\emph{Procceedings of the First Annual Conference of the International Technology Alliance in Network and Information Science}}.
\newblock


\bibitem[\protect\citeauthoryear{Tiferes and Bisantz}{Tiferes and Bisantz}{2018}]%
        {tiferes2018impact}
\bibfield{author}{\bibinfo{person}{Judith Tiferes} {and} \bibinfo{person}{Ann~M Bisantz}.} \bibinfo{year}{2018}\natexlab{}.
\newblock \showarticletitle{The impact of team characteristics and context on team communication: An integrative literature review}.
\newblock \bibinfo{journal}{\emph{Applied Ergonomics}}  \bibinfo{volume}{68} (\bibinfo{year}{2018}), \bibinfo{pages}{146--159}.
\newblock


\bibitem[\protect\citeauthoryear{Wang, Ma, Feng, Zhang, Yang, Zhang, Chen, Tang, Chen, Lin, et~al\mbox{.}}{Wang et~al\mbox{.}}{2024}]%
        {wang2024survey}
\bibfield{author}{\bibinfo{person}{Lei Wang}, \bibinfo{person}{Chen Ma}, \bibinfo{person}{Xueyang Feng}, \bibinfo{person}{Zeyu Zhang}, \bibinfo{person}{Hao Yang}, \bibinfo{person}{Jingsen Zhang}, \bibinfo{person}{Zhiyuan Chen}, \bibinfo{person}{Jiakai Tang}, \bibinfo{person}{Xu Chen}, \bibinfo{person}{Yankai Lin}, {et~al\mbox{.}}} \bibinfo{year}{2024}\natexlab{}.
\newblock \showarticletitle{A survey on large language model based autonomous agents}.
\newblock \bibinfo{journal}{\emph{Frontiers of Computer Science}} \bibinfo{volume}{18}, \bibinfo{number}{6} (\bibinfo{year}{2024}), \bibinfo{pages}{186345}.
\newblock


\bibitem[\protect\citeauthoryear{Webber, Detjen, MacLean, and Thomas}{Webber et~al\mbox{.}}{2019}]%
        {webber2019team}
\bibfield{author}{\bibinfo{person}{Sheila~Simsarian Webber}, \bibinfo{person}{Jodi Detjen}, \bibinfo{person}{Tammy~L MacLean}, {and} \bibinfo{person}{Dominic Thomas}.} \bibinfo{year}{2019}\natexlab{}.
\newblock \showarticletitle{Team challenges: Is artificial intelligence the solution?}
\newblock \bibinfo{journal}{\emph{Business Horizons}} \bibinfo{volume}{62}, \bibinfo{number}{6} (\bibinfo{year}{2019}), \bibinfo{pages}{741--750}.
\newblock


\bibitem[\protect\citeauthoryear{Zhang, McNeese, Freeman, and Musick}{Zhang et~al\mbox{.}}{2021}]%
        {zhang2021ideal}
\bibfield{author}{\bibinfo{person}{Rui Zhang}, \bibinfo{person}{Nathan~J McNeese}, \bibinfo{person}{Guo Freeman}, {and} \bibinfo{person}{Geoff Musick}.} \bibinfo{year}{2021}\natexlab{}.
\newblock \showarticletitle{" An ideal human" expectations of AI teammates in human-AI teaming}.
\newblock \bibinfo{journal}{\emph{Proceedings of the ACM on Human-Computer Interaction}} \bibinfo{volume}{4}, \bibinfo{number}{CSCW3} (\bibinfo{year}{2021}), \bibinfo{pages}{1--25}.
\newblock


\end{thebibliography}


\end{document}